\pdfoutput=1

\documentclass[11pt]{article}

\usepackage{naacl2021}
\usepackage{times}
\usepackage{latexsym}

\usepackage[T1]{fontenc}

\usepackage[utf8]{inputenc}

\usepackage{microtype}

\usepackage{graphicx}
\usepackage{amsmath}
\usepackage{subcaption}
\usepackage{array}
\usepackage{booktabs}
\usepackage{multirow}
\title{Integrating Regular Expressions with Neural Networks via DFA}

\author{Shaobo Li$^{1}$, Qun Liu$^{2}$, Xin Jiang$^{2}$, Yichun Yin$^{2}$
\\
\textbf{Chengjie Sun$^{1}$, Bingquan Liu$^{1}$, Zhenzhou Ji$^{1}$ and Lifeng Shang$^{2}$}\\
$^{1}$Harbin Institute of Technology\\
$^{2}$Huawei Noah's Ark Lab\\
\texttt{shli@insun.hit.edu.cn},
\texttt{\{sunchengjie, liubq, jizhenzhou\}@hit.edu.cn}\\
\texttt{\{qun.liu, Jiang.Xin, yinyichun, shang.lifeng\}@huawei.com}\\
}
\begin{document}
\maketitle
\begin{abstract}
Human-designed rules are widely used to build industry applications. However, it is infeasible to maintain thousands of such hand-crafted rules. So it is very important to integrate the rule knowledge into neural networks to build a hybrid model that achieves better performance. Specifically, the human-designed rules are formulated as Regular Expressions (REs), from which the equivalent Minimal Deterministic Finite Automatons (MDFAs) are constructed. We propose to use the MDFA as an intermediate model to capture the matched RE patterns as rule-based features for each input sentence and introduce these additional features into neural networks. We evaluate the proposed method on the ATIS intent classification task. The experiment results show that the proposed method achieves the best performance compared to neural networks and four other methods that combine REs and neural networks when the training dataset is relatively small.
\end{abstract}

\section{Introduction}
Although Neural Network~(NN) based approaches have been widely used in various natural language processing tasks and achieved remarkable results \cite{young2018recent}, there are still limitations of NNs faced by the community, such as the data-hungry nature, lacking interpretability ability and vulnerability to adversarial attacks. In most cases, the NN-based models cannot be directly applied to the scenario where there are limited training samples, where the rule-based methods can still work properly and are widely used. Integrating human-designed symbolic knowledge into NN-based models is believed to be a promising and practical way to alleviate these limitations \cite{garcez2019neural}, and many works have been explored in this direction \cite{liang2017neural,xie2019embedding,arora2020iterefine,luo2018marrying}. However, it is still an open problem to effectively integrate the highly abstract human knowledge encoded by discrete rules with the data-driven neural models. 

In this paper, we propose a novel method that adopts the regular expression~(RE) as human-designed rules and combines it with the NN-based model (i.e., NN-based sentence classifier). The key problem is {\it how to effectively represent the regular expression into numeric features}, and then {\it integrate them with neural models}. In conventional rule-based representations~\cite{luo2018marrying}, the final matching results of input sentences to REs~(i.e., accept or reject) or the occurrences of some key patterns in REs are directly used as the additional features. 

In this work, REs are converted into MDFAs, and MDFAs are used as intermediate models to capture more fine-grained MDFA-based features from the input sentence. The MDFA-based features easily integrate with neural networks and can be absorbed effectively. Furthermore, we propose two RE-NN hybrid models that incorporate the MDFA-based features into NN at different levels. Comprehensive experiments are conducted on ATIS intent classification dataset \cite{hemphill1990atis} with training sets of various sizes. The experiment results demonstrate that the proposed hybrid models improve the performance of NN and also perform better than the existing models which combine NNs with REs. 

\section{Method}

\begin{figure}[t]
  \vspace{-10pt}
  \setlength{\abovecaptionskip}{0pt}
   \setlength{\belowcaptionskip}{-10pt}
  \includegraphics[width=\columnwidth,clip,trim={18pt 4pt 35pt 7pt}]{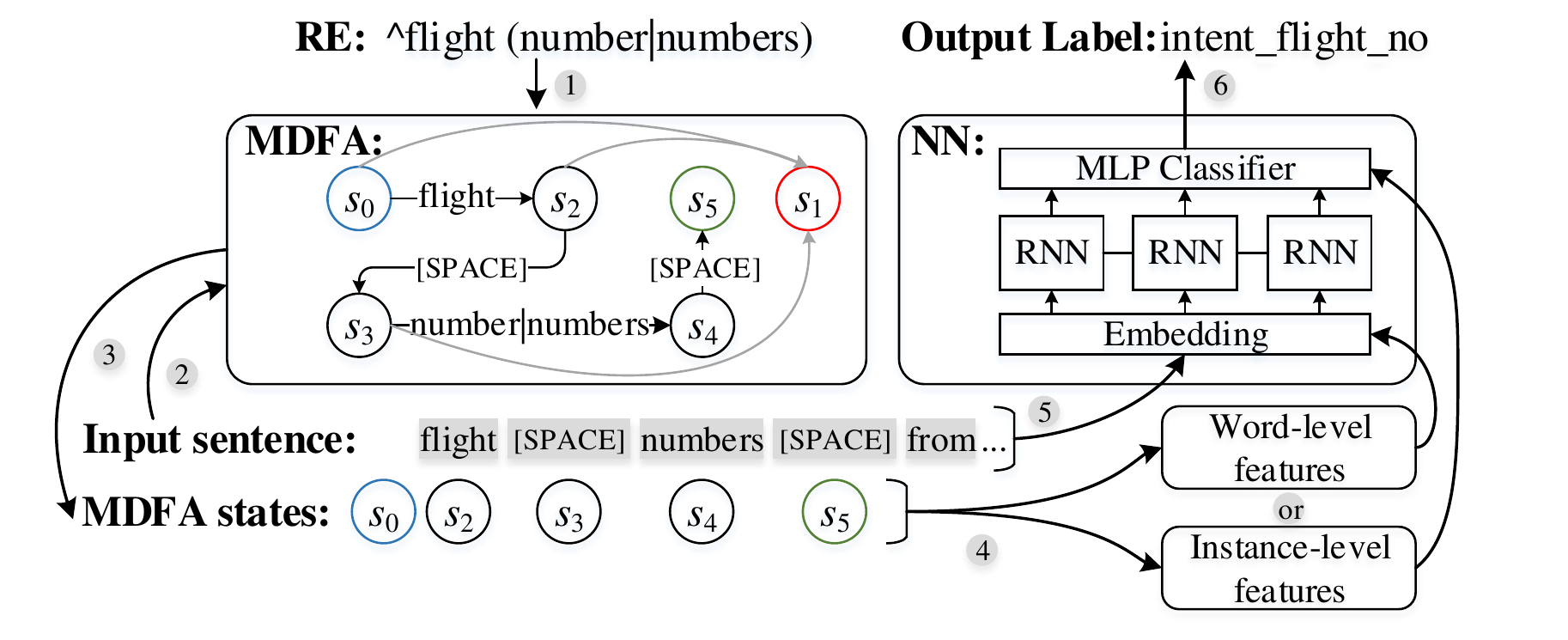}
  \centering
  \caption{The proposed method integrates REs with NN through MDFA.}
  \vspace{-8pt}
  \label{fig:overview}
\end{figure}

Figure~\ref{fig:overview} shows the proposed method. Each RE is first turned into MDFA with the algorithms in \cite{hopcroft2001introduction}. Based on the obtained MDFA, a sequence of transition states can be obtained for each input sentence, which presents local matching patterns defined by the RE occurring in the sentence. Then we transform this sentence-dependent state sequence into MDFA-based features and integrate them into NN as additional input at the instance level or word level. For the NN-based component, Recurrent Neural Network (RNN) \cite{jain1999recurrent} is used as the backbone.

The instance-level integration aims to encourage the MDFA-based features to interact with high-level NN-based representations of an entire sentence. Differently, the word-level integration associates the MDFA-based features with smaller-granularity semantic, i.e., word embeddings \cite{mikolov2013distributed}. The word-level integration strategy adopts MDFA-based features to guide and regularize RNN internal representations word by word, which is motivated by the recent findings \cite{michalenko2018representing} that there exists a strong structural relationship between internal representations of RNNs and MDFAs when recognizing formal languages. 

\subsection{Encode MDFA State Sequences into Numeric MDFA-based Features}
\label{sec:rule_encoding}
In this section, we describe how to encode MDFA state sequences into numeric features that contain the sentence-dependent information from rules. A regular expression (RE) can be transformed into an equivalent MDFA that embodies the human-designed rules in RE~\cite{hopcroft2001introduction}. Formally, an MDFA can be represented as $M=\{\mathcal{S},\mathcal{E},\delta,s_0,\mathcal{F}\}$, which contains a state set $\mathcal{S}$, an input symbol set $\mathcal{E}$, a transition function $\delta:\mathcal{S}\times \mathcal{E}\rightarrow \mathcal{S}$, an initial state $s_0\in \mathcal{S}$ and an end state set $\mathcal{F} \subset \mathcal{S}$. 

Matching RE with a sentence can be converted to a sequence of MDFA state transition. A sentence $W=\{w_1,w_2,\ldots,w_n\}$ that contains $n$ words is fed into the MDFA word by word. Start from the initial state $s_0$, the MDFA will transfer from the current state to a target state depending on the current received word and the transition function $\delta$. In this way, a sequence of target state $S$ can be observed sequentially until one end state is reached or all the words in $W$ are exhausted. If $S$ ends with an end state in $\mathcal{F}$, it means that the RE (or equivalently, the MDFA) accepts the sentence $W$\footnote{The conventional RE is character-based and results in a character-triggered MDFA that contains massive fine-grained states. To reduce the number of states and make these states more informative, we employ the word-based RE instead of character-based by treating a whole word as a single symbol.}.

The state sequence $S$ contains a detailed procedure about how a RE examines the sentence. We encode $S$ as MDFA-based features for the sentence. Suppose there are $p$ REs for different labels, we can obtain $p$ MDFA state sequences by feeding $W$ into each RE. We use superscript $k$ to identify the $k$-th RE and represent the MDFA state sequence produced by the $k$-th as $S^k$. 

\begin{figure}[t]
  \vspace{-5pt}
   \setlength{\abovecaptionskip}{0pt}
   \setlength{\belowcaptionskip}{-10pt}
  \includegraphics[width=\columnwidth,clip,trim={0pt 6pt 0pt 1pt}]{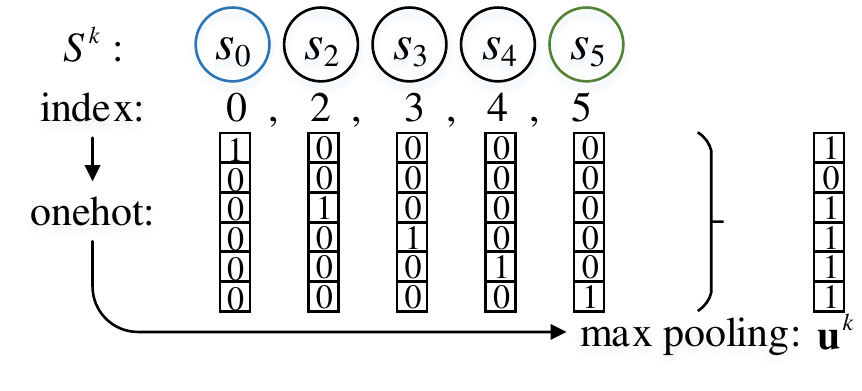}
  \centering
  \caption{Encode the MDFA state sequence as an instance-level feature vector.}
  \label{fig:instance_level}
  \vspace{-5pt}
\end{figure}

We propose two types of encoding, named instance-level encoding and word-level encoding, which encode MDFA state sequences as instance-level features or word-level features respectively. The instance-level encoding is to encode each MDFA state sequence into a single feature vector. Specifically, for the $k$-th RE, all the different states in the state set of the constructed MDFA are indexed with consecutive integers. Then the $S^k$ is converted into a one-hot vector sequence by converting each state in $S^k$ by a one-hot vector based on the integral index of it. Finally, the one-hot vector sequence is aggregated into a single vector $\mathbf{u}^k$ via max-pooling operation. Figure~\ref{fig:instance_level} illustrates an example of calculating $\mathbf{u}^k$. 

The word-level encoding encodes the MDFA state sequences as word-level features, by which every word in $W$ is assigned with binary tags from different REs. When matching the $k$-th RE with $W$, the state transition is triggered word by word by feeding the words in $W$ into MDFA orderly. We denote the state arrived at after MDFA consuming the $i$-th word $w_i$ in $W$ as $s_{w_i}^k$ and align it with $w_i$. Inspired by the BIO (Beginning-Inside-Outside) tagging format~\cite{ramshaw1999text}, the state $s_{w_i}^k$ is simplified to a binary tag $v_{w_i}$ for $w_i$ as:
\begin{itemize}
  \item $w_i$ is tagged with 1 ($v_{w_i}=1$) if the $k$-th RE accepts $W$ and $s_{w_i}^k\in S^k$, which means the $w_i$ is inside of the matching procedure and results in acceptance. 
  \item $w_i$ is tagged with 0 ($v_{w_i}=0$) otherwise.
\end{itemize}
Because of the high-precision low-recall nature of human-designed rules, the ``accept'' output by RE is more believable than the ``reject''. Therefore, the MDFA-based encoding is only activated when the input sentence is accepted by the RE, set to zero otherwise. In this way, binary tag sequence $v^k=\{v_{w_1}^k,v_{w_2}^k,\ldots,v_{w_n}^k\}$ can be obtained for $W$ based on the $k$-th RE. 

Finally, we can obtain two types of MDFA-based numeric features, $\{\mathbf{u}^k\}$ and $\{v^k\}$ ($k={1,\ldots,p}$), which will be further incorporated into the NN-based component at the instance level or word level respectively.

\subsection{NN-based Component}
We use a pure NN-based sentence classifier, that consists of a BLSTM (Bidirectional LSTM) with attention mechanism as feature extractor \cite{zhou2016attention} and a MLP (Multi-Layer Perceptron) as classifier \cite{amendolia2003comparative}, as the NN-based component in the proposed hybrid models. Specifically, for the input sentence $W=\{w_1,w_2,\ldots,w_n\}$, BLSTM first generates a hidden state sequence $\mathbf{H}=\{\mathbf{h}_1,\mathbf{h}_2,\ldots,\mathbf{h}_n\}$ as
\begin{equation}
  \mathbf{H}={\rm BLSTM}(\{\mathbf{e}(w_1),\ldots,\mathbf{e}(w_n)\}),
  \label{eq:blstm}
\end{equation}
where $\mathbf{e}(\cdot)$ denotes the embedding vector of the input word. $\mathbf{H}$ is transformed to a fixed-length feature vector $\mathbf{f}$ through attention mechanism. Then $\mathbf{f}$ is fed to the MLP classifier to calculate the class probabilities $\mathbf{y}$:
\begin{equation}
  \alpha_{i}=\frac{\exp(\mathbf{h}_i\mathbf{W}\mathbf{h}_n)}{\sum_{j}{\exp(\mathbf{h}_j\mathbf{W}\mathbf{h}_n)}},
  \enspace\enspace
  \mathbf{f}=\sum_{i=1}^{n}{\alpha_{i}}\mathbf{h}_{i}
\end{equation}
\begin{equation}
  \mathbf{y}=
    {\rm MLP}(\mathbf{f})
\label{eq:mlp}
\end{equation}
This NN-based component is also used as our NN baseline, referred to as \textsc{nnsc}. 

\subsection{Introduce the MDFA-Based Features into the NN-based Component}
We import two types of MDFA-based features into the NN-based component at the instance or word level to build two hybrid models. In the first hybrid model, referred to as \textsc{instance}, $\{\mathbf{u}^k\}$ is introduced into MLP classifier companying with the NN-based feature $\mathbf{f}$, which aims to enrich the NN-based representation of the entire input sentence with instance-level features. This hybrid model calculates the probabilities by 
\begin{equation}
  \mathbf{y}=
    {\rm MLP}
    ([\mathbf{f};\mathbf{u}^1;\ldots;\mathbf{u}^p])
\end{equation}
instead of Equation~\eqref{eq:mlp} in the \textsc{nnsc}. $[;]$ denotes concatenating of vectors.



The other hybrid model named \textsc{word} incorporates the word-level features $\{v^k\}$ into word embedding. $p$ REs produce $p$ binary tag sequences $\{v^k\}$ and each word in the input sentence corresponds to $p$ binary tags. All these $p$ binary tags are appended to the corresponding word embedding vector. For word $w_i$ in the input sentence $W$, the word embedding $\mathbf{e}(w_i)$ is expanded to MDFA-enhanced word embedding ${\mathbf{e}}_v(w_i)$:
\begin{equation}
  \mathbf{e}_v(w_i)=[\mathbf{e}(w_{i});v_{w_i}^1;\ldots;v_{w_i}^p]
\end{equation}
And Equation~\eqref{eq:blstm} in \textsc{nnsc} is modified to 
\begin{equation}
  \mathbf{H}={\rm BLSTM}(\{\mathbf{e}_v(w_1),\ldots,\mathbf{e}_v(w_n)\})
\end{equation}
in this hybrid model when calculating $\mathbf{y}$. 

  

\section{Experiments}

\begin{figure*}[t]
  \vspace{-10pt}
   \setlength{\abovecaptionskip}{1pt}
   \setlength{\belowcaptionskip}{-14pt}  
  \includegraphics[width=\textwidth,clip,trim={0pt 10pt 0pt 0pt}]{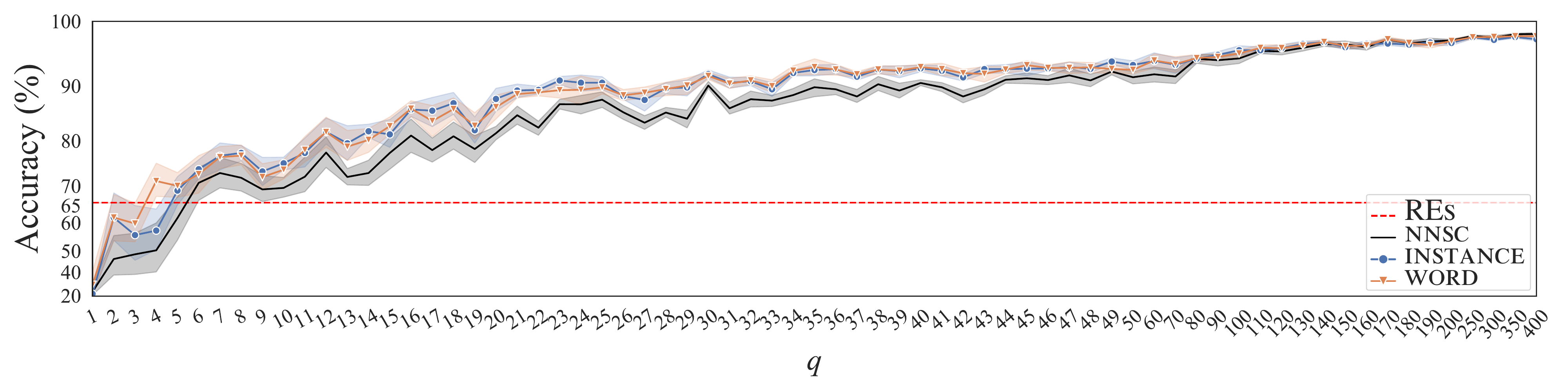}
  \centering
  \caption{Comparisons of the REs, NN-based model (denoted as \textsc{nnsc}) and the proposed hybrid models (denoted as \textsc{instance} and \textsc{word}).}
  \label{fig:acc}
\end{figure*}

\subsection{Settings}
We evaluate the models in few-shot settings, where the training samples are insufficient to training the NN-based model to achieve commendable performance, to see if the hybrid model boosts the performance by introducing human-designed rules. To further show the effectiveness of the proposed MDFA-based integration, we compared our hybrid models with other hybrid models that combine REs with NN without using MDFA. 

We employ the widely used ATIS intent classification dataset\footnote{https://pfllo.github.io/data/ACL18-data\_split.zip}. This dataset contains 4,978 training samples and 893 test samples with 18 intent labels. There are 54 manually written REs for intent classification obtained from \citet{luo2018marrying}\footnote{https://pfllo.github.io/data/ACL18-REs.zip}. We build few-shot training sets by randomly selecting $q$ samples for each class from the full training set. The models are trained on each training set respectively and evaluated on the official test set.

\subsection{Comparison with NN and Other RE-NN Hybrid Models}

\begin{table}[htbp]
  \small
  \centering
  \begin{tabular}{p{0.25\columnwidth}|>{\centering}p{0.05\columnwidth} >{\centering}p{0.05\columnwidth} >{\centering}p{0.06\columnwidth} >{\centering}p{0.075\columnwidth} >{\centering}p{0.075\columnwidth} p{0.08\columnwidth}<{\centering}}
  \toprule
  \multirow{2}{*}{Model}            & \multirow{2}{*}{$q$=5}     & \multirow{2}{*}{$q$=10}    & \multirow{2}{*}{$q$=20}    & $q$=5       & $q$=10       & $q$=20 \\
                                    &           &           &           & w/300       & w/300        & w/300  \\
  \midrule
  \textsc{mlp-i}*                   & 63.72     & 73.46     & 83.20     & 94.23       & 94.90        & 95.71     \\
  \textsc{mlp-o}*                   & 58.68     & 77.83     & 89.25     & 91.82       & 97.09        & 97.15     \\
  \textsc{attn} \tiny{(luo et al.)} & 75.36     & 85.44     & 88.80     & 92.05       & 96.98        & 97.76     \\
  \textsc{fol}  \tiny{(hu et al.)}  & 56.22     & 68.42     & 84.10     & 91.94       & 96.75        & 97.42     \\
  \midrule
  \textsc{nnsc} \tiny{(ours)}       & 65.28     & 73.90     & 86.89     & 94.06       & 96.97        & 98.20     \\
  \textsc{instance} \tiny{(ours)}   & \bf 81.97 & 84.99     & \bf 91.04 & 94.51       & \bf 97.64    & 97.87     \\
  \textsc{word} \tiny{(ours)}       & 80.17     & \bf 86.67 & 90.25     & \bf 94.62   & 97.20        & \bf 98.32 \\
  \bottomrule
  \end{tabular}
  \caption{\label{tab:marry_comp} Comparisons with the other methods that combine REs with NN without MDFA and the NN-based Model (denoted as \textsc{nnsc}). * indicates the baseline hybrid models described in \cite{luo2018marrying}. The results are reported in accuracy.
   }
  \vspace{-10pt}
\end{table}
Our hybrid models are compared with the pure NN-based model (denoted as \textsc{nnsc}) and four RE-NN hybrid models that introduce REs into NN without using MDFA, as shown in Table~\ref{tab:marry_comp}\footnote{``w/300'' represents 300 extra training samples from each top 3 frequency classes are used additionally. These training sets are used in \cite{luo2018marrying} and imported for fair comparisons.}. 
\textsc{mlp-i} feeds the final matching results of REs into the MLP classifier as additional features, while hybrid model \textsc{instance} introduces the more fine-grained MDFA-based encodings. \textsc{mlp-o} add the RE matching result to the probabilities output by the MLP classifier, \textsc{attn} uses the keywords in REs to regularize the attention weights in \textsc{nnsc}. \textsc{fol} converts the REs into FOL (First-Order-Logic) rules and distills the knowledge from FOL rules into NN using the method proposed by \citep{hu2016harnessing}.

It can be seen that \textsc{instance} and \textsc{word} achieve better performances. We conjecture that this is because the MDFA-based features can capture more fine-grained information from REs, and these features can be well absorbed by neural networks with the proposed integration strategies. 

\subsection{Experiments on Training Sets of Different Sizes}
To further investigate the performance of the models on training sets of different sizes, we selected 69 values for $q$ from 1 to 400 and constructed the training sets accordingly. To reduce the impact of randomness, we repeat each selection three times and the experiment on each training set 5 times. The performances of the proposed models and \textsc{nnsc} are graphically shown in Figure~\ref{fig:acc}, with the shadow representing the 95\% confidence interval. 

Note that the proposed hybrid models achieve significant improvements when the amount of training data is small (e.g., $q\leq 30$). Due to the elements in the MDFA-based features has a small value set (i.e., 0, 1), such features have fewer combinations than natural language constituted by thousands of different words. Therefore, the MDFA-based features can be learned even with fewer training samples. However, in the case when the training dataset is large enough, the REs (i.e., 65.7\% accuracy shown as the dashed line in Figure~\ref{fig:acc}) are much weaker than \textsc{nnsc}. The performance is therefore hard to be boosted by additionally utilizing rules, 
and the improvements over \textsc{nnsc} decrease as the training set size increases. 

The average accuracy improvement over all the constructed training sets that the proposed \textsc{instance} and \textsc{word} models achieved are 2.55\% and 3.99\%, respectively. The word-level integration performs relatively better than the instance-level. We think it is because there exist rich and flexible interactions between RNN and rule-based features. 

\section{Conclusions}
In this paper, we tried to incorporate human-designed REs into NN-based models. The REs are transformed into MDFAs to provide MDFA-based features that can be easily absorbed by NN. Moreover, we proposed two kinds of hybrid models that incorporate the MDFA-based features into NN at word level or instance level. The experiment results on the ATIS intent classification task have shown that the proposed hybrid models improve the performance of NN when the training data is insufficient and further outperform the other RE-NN hybrid models without using MDFA.

\bibliography{anthology,custom}
\bibliographystyle{acl_natbib}




\end{document}